\pdfoutput=1

\documentclass[11pt]{article}

\usepackage{acl}

\usepackage{times}
\usepackage{latexsym}

\usepackage[T1]{fontenc}

\usepackage[utf8]{inputenc}

\usepackage{microtype}

\usepackage{inconsolata}

\usepackage{graphicx}

\usepackage{textcomp}
\usepackage{amssymb}
\usepackage{multirow}
\usepackage{amsmath}
\usepackage{booktabs}
\usepackage{pifont}
\usepackage{color}
\usepackage{bm}
\usepackage{answers}
\usepackage{xcolor}
\usepackage{algorithm}
\usepackage{algorithmic}
\usepackage{autobreak}

\title{Self-Evolutionary Large Language Models through Uncertainty-Enhanced Preference Optimization}

\author{
Jianing Wang$^{1}$\thanks{~~J. Wang obtained the Ph.D. degree at East China Normal University.}, Yang Zhou$^{1}$, Xiaocheng Zhang$^{1,2}$, Mengjiao Bao$^{1}$, Peng Yan$^{1}$\thanks{~~Corresponding author.} \\
$^{1}$ Meituan, $^{2}$ Harbin Institute of Technology \\
\texttt{\{wangjianing16, yanpeng04\}@meituan.com}
}

\begin{document}

\maketitle

\begin{abstract}

Iterative preference optimization has recently become one of the de-facto training paradigms for large language models (LLMs), but the performance is still underwhelming due to too much noisy preference data yielded in the loop.
To combat this issue, we present an \textbf{U}ncertainty-enhanced \textbf{P}reference \textbf{O}ptimization (UPO) framework to make the LLM self-evolve with reliable feedback.
The key idea is mitigating the noisy preference data derived from the current policy and reward models by performing pair-wise uncertainty estimation and judiciously reliable feedback sampling.
To reach this goal, we thus introduce an estimator model, which incorporates Monte Carlo (MC) dropout in Bayesian neural network (BNN) to perform uncertainty estimation for the preference data derived from the LLM policy.
Compared to the existing methods that directly filter generated responses based on the reward score, the estimator focuses on the model uncertainty in a pair-wise manner and effectively bypasses the confirmation bias problem of the reward model.
Additionally, we also propose an uncertainty-enhanced self-evolution algorithm to improve the robustness of preference optimization and encourage the LLM to generate responses with both high reward and certainty.
Extensive experiments over multiple benchmarks demonstrate that our framework substantially alleviates the noisy problem and improves the performance of iterative preference optimization
~\footnote{The code will be released at \url{https://github.com/wjn1996/Uncertainty-Preference-Optimization}.}.
\end{abstract}

%

\section{Introduction}
Recently, the NLP community has witnessed the success of preference optimization for large language models (LLMs), which has become one of the significant ingredients of recent revolutions~\cite{Brown2020Language, OpenAI2023GPT4, Tunstall2023Zephyr, Zheng2023Secrets}.
As a post-training process of LLM, preference optimization aims to align the LLM policy with the labeled human feedback or AI feedback data.
Early approaches utilize reinforcement learning (RL) to train the LLM policy online based on the human feedback simulated by a tuned reward model, referred to as RLHF~\cite{Christiano2017Deep, Lee2021PEBBLE, Ouyang2022Training}.
Besides, offline direct preference optimization (DPO) and some variants view LLM-as-judge~\cite{Yuan2024Self} and directly align the policy with feedback~\cite{Rafailov2023Direct, Ethayarajh2024KTO}.

\begin{figure}[t]
\centering
\includegraphics[width=\linewidth]{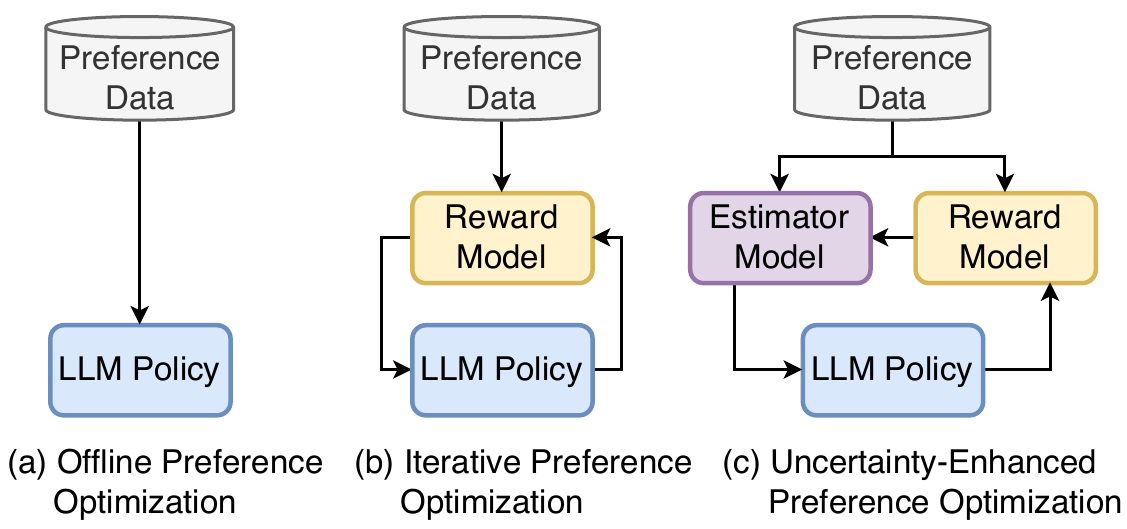} 
\caption{Overview of three paradigms.}
\label{fig:paradigm}
\end{figure}

Despite the success, these approaches relied on massive labeled preference data which requires tons of manpower and resources. 
To combat this issue, some recent researches introduce a novel iterative preference optimization~\cite{Pang2024Iterative, Chen2024Self, Kim2024Aligning, Xu2023Some, Rosset2024Direct, Wu2024Self, Xie2024Monte}. 
As shown in Figure~\ref{fig:paradigm} (b), the offline methods can be iteratively applied similarly to the self-training procedure, where the previously trained policy generates new preference data which are then used to train the new policy.
Generally, a reward model is also required in the iteration to simulate feedback for self-evolve~\cite{Xu2024WizardLM, Tao2024A}.

However, we find one of the potential pitfalls in the iteration is that the reward model may assign unsuitable scores for the responses, leading to deriving multiple noisy preference pairs and hindering performance. This problem gets exaggerated when the interaction number increases~\cite{Han2018Co, Choi2024Self}.
Hence, the paramount challenge is meticulously selecting reliable preference data and making the preference optimization not distorted by noise.
A simple solution is to choose one pair in which two responses ignifying a notable disparity in terms of the reward score~\cite{Pang2024Iterative}. Yet, it can not bypass the confirmation bias problem~\cite{Andersen2022Efficient, Rizve2021In, Wang2021Combating} in the self-training-like paradigm. 

To this end, we present an \textbf{U}ncertainty-enhanced \textbf{P}reference \textbf{O}ptimization (UPO) framework to circumvent the noise problem.
To elaborate, we introduce an estimator model that essentially performs a classification task to detect which response is more suitable for the query. 
As shown in Figure~\ref{fig:paradigm} (c), different from the existing reward model that can only assign a scalar score in the inference stage, 
it can be equipped with a Monte Carlo (MC) dropout technique, which is the approximation technique in Bayesian Neural Network (BNN)~\cite{Gal2016Dropout, Wang2016Towards}, to estimate the uncertainty of each preference pair.
Thus, a sampling signal based on the model certainty can be used to represent the reliability of the preference pair.
To further improve the robustness of the iteration preference optimization, we additionally develop an uncertainty-enhanced self-evolution algorithm. 
Specifically, we first use the estimator certainty to split the generated preference data into reliable pairs and unreliable pairs, where reliable pairs can easily provide high-quality feedback and unreliable pairs are quite hard to express the preference.
We thus integrate the uncertainty into DPO to encourage the LLM policy to know what generated pairs are reliable or unreliable feedback.
Therefore, with the dual blessing of rewards and uncertainty, the new LLM policy can generate responses with both high rewards and high certainty.

We conduct extensive experiments on two universal NLP benchmarks (i.e., AlpacaEval 2.0~\cite{Dubois2024Length} and MT-Bench~\cite{Zheng2023Judging}) and two mathematics reasoning tasks (i.e., GSM8K~\cite{cobbe2021training} and MATH~\cite{Hendrycks2021Measuring}), results demonstrate that our UPO framework substantially enhances the effectiveness of preference alignment, and achieves the best performance in auto evaluation.



\section{Preliminaries}

We first introduce the background knowledge of the iteration preference optimization and Bayesian neural network.

\subsection{Preference Optimization}

Suppose that the LLM policy is denoted as $\pi_{\theta}$ and it has been tuned after the pre-training and supervised fine-tuning (SFT) stage.
The goal of preference optimization is to post-train the LLM policy on well-manual preference data.
Formally, given a labeled preference data $\mathcal{D}=\{(x, y_{w}, y_{l})\}$ which consists of multiple triples~\footnote{In this paper, $(x, y_w, y_l)$ is named as preference triple or preference data, while $(y_w, y_l)$ is named as preference pair.} conditioned by a prompt $x\in\mathcal{X}$, a preferred response $y_{w}\in\mathcal{Y}$ as the winner (chosen) and a dispreferred response $y_{l}\in\mathcal{Y}$ as the loser (rejected). $\mathcal{X}$ and $\mathcal{Y}$ are respectively prompt and output distributions.

During the optimization, a series of methods leverage RLHF to process the feedback online. 
Generally, it requires a reward model pre-trained on the preference data through the Bradley-Terry model~\cite{bradley1952rank} as:
\begin{equation}
\begin{aligned}
p(y_w\succ y_l)=\frac{\exp{(r_{\phi}(x, y_w))}}{\exp{(r_{\phi}(x, y_w))} + \exp{(r_{\phi}(x, y_l))}},
\label{eql:bradley-terry}
\end{aligned}
\end{equation}
where $r_{\phi}(x, y)$ is the reward model and outputs a scaler score as the reward of response $y$ towards the given prompt $x$.
The parameters of $r_{\phi}(x, y)$ can be updated as the following maximum-likelihood objective:
\begin{equation}
\begin{aligned}
\mathcal{L}_{r}(\phi) =& -\mathbb{E}_{(x, y_w, y_l)\sim\mathcal{D}} \\
& [\log\sigma(r_{\phi}(x, y_w) - \sigma(r_{\phi}(x, y_l)))],
\label{eql:reward-loss}
\end{aligned}
\end{equation}
where $\sigma(\cdot)$ is the sigmoid function. When a pre-trained reward model is available, the LLM policy can be repetitively aligned to the new pairs derived from the reward model with a proximal policy optimization (PPO) algorithm:
\begin{equation}
\begin{aligned}
\mathcal{L}_{\text{rlhf}}(\theta) = &-\mathbb{E}_{x\sim\mathcal{X}, y\sim\pi_{\theta}(\cdot|x)}[r_{\phi}(x, y)] \\
&+ \beta\mathbb{E}_{x\sim\mathcal{X}}[\text{KL}(\pi_{\theta}(\cdot|x)||\pi_{\text{ref}}(\cdot|x))],
\label{eql:ppo-loss}
\end{aligned}
\end{equation}
where $\beta>0$ is the balance factor, the KL divergence $\text{KL}(\cdot||\cdot)$ aims to maintain the original output distribution similar to the consistency regularization. $\pi_{\text{ref}}$ is the reference model which shares the same parameters with $\pi_{\theta}$ but is frozen after the SFT stage.

In contrast to RLHF, DPO aims to follow the LLM-as-judge paradigm by directly optimizing the policy:
\begin{equation}
\begin{aligned}
\mathcal{L_{\text{dpo}}}(\theta) = -\mathbb{E}_{(x, y_w, y_l)\sim\mathcal{D}} \log\sigma(\beta h_{\pi_{\text{ref}}}^{\pi_{\theta}}(x, y_w, y_l)),
\label{eql:dpo-loss}
\end{aligned}
\end{equation}
where $h_{\pi_{\text{ref}}}^{\pi_{\theta}}(x, y_w, y_l)$ is the reward difference between prefered response and disprefered response:
\begin{equation}
\begin{aligned}
h_{\pi_{\text{ref}}}^{\pi_{\theta}}(x, y_w, y_l) = \log\frac{\pi_{\theta}(y_w|x)}{\pi_{\text{ref}}(y_w|x)} - \log\frac{\pi_{\theta}(y_l|x)}{\pi_{\text{ref}}(y_l|x)}.
\label{eql:dpo-reward-margin}
\end{aligned}
\end{equation}

\begin{figure*}[t]
\centering
\includegraphics[width=\textwidth]{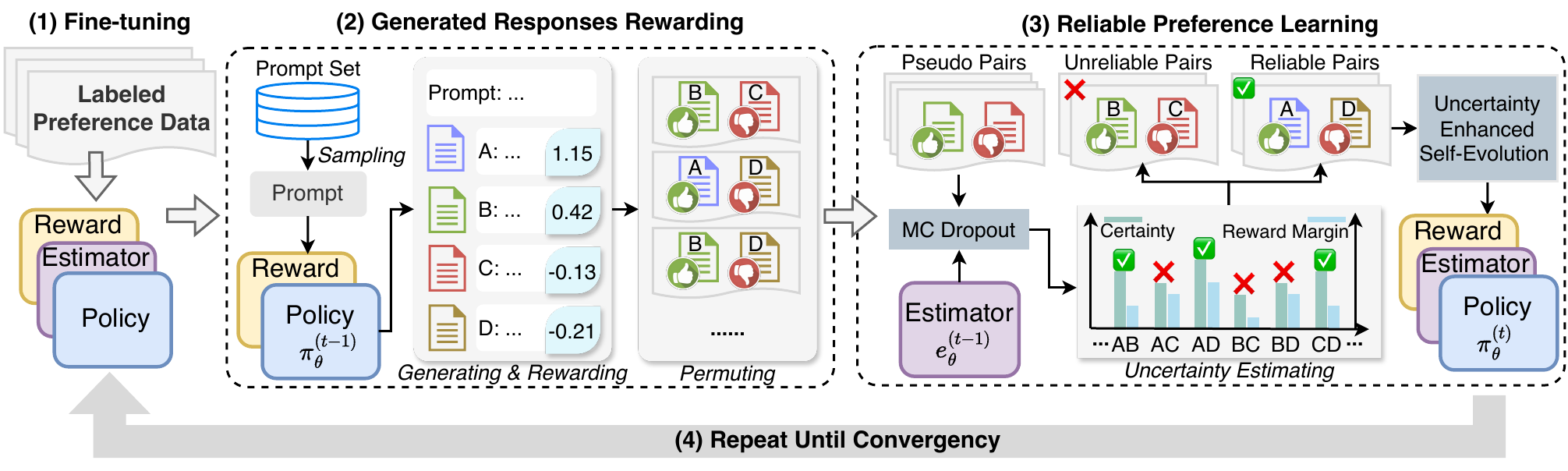} 
\caption{Illustration of UPO framework. We first use the labeled preference data to train a LLM policy, a reward model, and an estimator model. Then, multiple new preference data can be generated by the LLM policy based on the reward score. Finally, we use the uncertainty estimation technique to sample reliable data and further update the LLM policy with an uncertainty-enhanced self-evolution algorithm. The whole procedure repeats until convergence.}
\label{fig:framework}
\end{figure*}

\subsection{Bayesian Neural Network (BNN)}
In the iteration procedure, the preference pairs derived from the reward model or LLM itself may contain noisy data and hinder the whole performance.
We thus briefly describe the knowledge of BNN as the basic support for denoising.
Concretely, suppose a neural model $f_{\psi}$ can predict the preference, the vanilla BNN assumes a prior distribution over its model parameters $\psi$. 
In other words, BNN averages over all the possible weights instead of directly optimizing for the weights~\cite{Mukherjee2020Uncertainty}. 
Given a labeled preference $\mathcal{D}$, the parameter can be optimized by the posterior distribution $p(\psi|\mathcal{D})$. 
During model inference, given one unlabeled triple $(x, y_w, y_l)\in\mathcal{D}_{u}$ where $\mathcal{D}_{u}$ is the responses set generated by the LLM policy and reward model, the probability distribution can be formed as:
\begin{equation}
\begin{aligned}
p(c|x, y_w, y_l)=\int_{\psi}p(c|f_{\psi}(x, y_w, y_l)p(\psi|\mathcal{D}_{u})d\psi,
\label{eql:bnn-proba}
\end{aligned}
\end{equation}
where $c\in\{0, 1\}$ is the label represents $y_w\succ y_l$ is unsuitable or suitable.
To make the equation tractable, we can find a surrogate tractable distribution $q(\psi)$ based on a dropout distribution~\cite{Srivastava2014Dropout} that makes the model posterior easy to calculate. 
Thus, we can sample $T$ masked model weights $\{\widetilde{\psi}_t\}_{t=1}^{T}\sim q(\psi)$ from the current model. The approximate posterior is:
\begin{equation}
\begin{aligned}
p(c|x, y_1, y_2)\approx\frac{1}{T}\sum_{t=1}^{T}p(c|f_{\widetilde{\psi}_t}(x, y_1, y_2)).
\label{eql:bnn-posterior}
\end{aligned}
\end{equation}

\section{Methodology}
In this section, we develop an \textbf{U}ncertainty-enhanced \textbf{P}reference \textbf{O}ptimization (UPO) framework illustrated in Figure~\ref{fig:framework}, 
specialized for the improvement of the LLM self-evolve through iteration preference optimization paradigm.
The framework consists of three main procedures, i.e., initial stage fine-tuning, generated responses rewarding, and reliable preference learning.

\subsection{Initial Stage Fine-tuning}

In the initial stage, suppose that there is a supervised fine-tuned LLM $\pi_{\text{sft}}$ and a corresponding labeled preference data $\mathcal{D}^{(0)}$ derived from human or AI feedback.
We follow the previous works~\cite{Pang2024Iterative, Ouyang2022Training, Rafailov2023Direct, Kim2024Aligning} to use the initialized preference data to train a reward model $r_{\phi}^{(0)}$ based on the Bradley-Terry model in Eq.~\ref{eql:bradley-terry}, and a weak LLM policy $\pi_{\theta}^{(0)}$ optimized from $\pi_{\text{sft}}$ via DPO in Eq.~\ref{eql:dpo-loss}~\footnote{In fact, the reward model can be omitted when using DPO because the LLM policy can provide implicit rewards. Yet, we still train an explicit reward model which can be used freely in practical application.}.

In addition, we also develop an estimator which is essentially a binary classifier that detects whether a pair is suitable.
Different from the reward model that only assigns a scaler score, the estimator model can provide the probability of the fact that the preferred response is better than the dispreferred one, and will be used for uncertainty estimation in the reliable preference learning stage.
To train the model, we need to reform the existing preference data.

We first transform the original preference triple $(x, y_w, y_l)\in\mathcal{D}^{(0)}$ into a unified prompt, and the template is denoted as $\mathcal{T}(x, y_w, y_l)$ demonstrated in Appendix~\ref{sec:app-prompt}.
Therefore, we can construct a binary classification dataset to train an estimator model.
To make the training easier, we directly choose the backbone from $\pi_{\theta}^{(0)}$ and add an external classification head to project the last layer's representations at the last token position into a binary space. The training objective is formulated as:
\begin{equation}
\begin{aligned}
\mathcal{L}_{\text{est}}(\psi) = -\mathbb{E}_{(x, y_w, y_l)\sim\mathcal{D}^{(0)}}\log f_{\psi}(\mathcal{T}(x, y_w, y_l)).
\label{eql:estimator-loss}
\end{aligned}
\end{equation}


\subsection{Generated Responses Rewarding}

The LLM policy will be iteratively updated with the coordination of reward and estimator models.
For the $i$-th iteration, we assume that the current LLM policy is $\pi_{\theta}^{(i-1)}$.
In pursuit of obtaining more preference data to evolve the policy, we urge $\pi_{\theta}^{(i-1)}$ to generate multiple responses from new sampled prompts.
Specifically, give a prompt $x\in\mathcal{X}$, the corresponding responses can be represented as $\{y_j\}_{j=1}^{N}\sim\pi_{\theta}^{(i-1)}(\cdot|x)$, where $N\geq 4$ is the number of responses.
After that, the reward model $r_{\phi}^{(i-1)}$ at the previous stage will be used to assign a scale score for each response. Hence, we can sort the responses with the reward score and obtain all permutations.

Considering that too many permutations of each prompt will affect the execution efficiency of the framework, we pre-screen these permutations by a simple heuristic rule: we remove the pair whose chosen response (i.e., winner $y_w$) has a lower rank or rejected response (i.e., loser $y_l$) has a higher rank. For example, if we get six responses in descending sort (has a total of 15 pairs) and the top three responses are viewed as higher rank, only no more than 9 pairs will be used, expediting the process of iteration procedure because fewer data need to be estimated in the next stage.
At last, we denote the final generated permutations with the corresponding prompt as the pseudo preference pairs $\mathcal{D}_{u}^{(i)}$.

\subsection{Reliable Preference Learning}

In this stage, we aim to leverage the trained estimator model~\footnote{We do not directly leverage the probability from Eq.~\ref{eql:bradley-terry} because its objective is different from uncertainty estimation in BNN.} to select reliable reference data based on uncertainty estimation.

Given an estimator model $f_{\psi}^{(i-1)}$ and a pseudo preference data $\mathcal{D}_{u}^{(i)}$ generated by LLM policy and reward model.
We assume that each preference triple is independent of another and can be measured individually.
Specifically, we follow~\cite{Houlsby2011Bayesian, Wang2023Uncertainty} to leverage information gain of the model parameters to estimate how certain the estimator model is to the triple with respect to the true preference.
Therefore, we can obtain the formulation:
\begin{equation}
\begin{aligned}
\mathbb{B}(\tilde{c}_{j}, \psi|\mathcal{T}_{j}, \mathcal{D}_{u}^{(i)}) = 
& \mathbb{H}(\tilde{c}_{j}|\mathcal{T}_{j}, \mathcal{D}_{u}^{(i)}) - \\
& \mathbb{E}_{p(\psi|\mathcal{D}_{u}^{(i)})}[\mathbb{H}(\tilde{c}_{j}|\mathcal{T}_{j}, \psi)],
\label{eql:information-gain}
\end{aligned}
\end{equation}
where $\mathbb{H}(\cdot)$ is the entropy, 
$\mathcal{T}_{j}=\mathcal{T}(x_j, y_{wj}, y_{lj})$ is the input template of $j$-th triple from $\mathcal{D}_{u}^{(i)}$.
$\tilde{c}_{j}\in\{0, 1\}$ denote the prediction of estimator model. 
$p(\psi|\mathcal{D}_{u}^{(i)})$ is the posterior distribution.
Through this information gain, we can find that a lower $\mathbb{B}(\tilde{c}_{j}, \psi|\mathcal{T}_{j}, \mathcal{D}_{u}^{(i)})$ value means that the estimator model is more certain about the prediction, as higher certainty corresponds to lower information gain. In other words, the preference triples with higher certainty and is more reliable feedback towards the prompt.

For the implementation details, we use MC Dropout in BNN to estimate the information gain. Specifically, we open the dropout and repeat $T$ (default set as 10) times to get independent and identically distributed (i.i.d.) predictions:
\begin{equation}
\begin{aligned}
\hat{\mathbb{B}}(\tilde{c}_{j}, & \psi|\mathcal{T}_{j}, \mathcal{D}_{u}^{(i)}) = \\ 
&- \sum_{c\in\{0, 1\}}(\frac{1}{T}\sum_{t=1}^{T}\hat{p}_c^t)\log(\frac{1}{T}\sum_{t=1}^{T}\hat{p}_c^t) \\
&+ \frac{1}{T}\sum_{t=1}^{T}\sum_{c\in\{0, 1\}}\hat{p}_c^t\log(\hat{p}_c^t),
\label{eql:information-gain-mc}
\end{aligned}
\end{equation}
where $\hat{p}_c^t=p(c|f_{\widetilde{\psi}_t}(\mathcal{T}_{j}))$ is the predict probability for the triple $(x_j, y_{wj}, y_{lj})$ derived from the $t$-th masked model $\widetilde{\psi}_{t}\sim q(\psi)$.

\subsection{Uncertainty-Enhanced Self-Evolution}
In the reliable preference learning stage, we also present an uncertainty-enhanced self-evolution algorithm to improve the robustness of LLM alignment.
Based on the uncertainty estimation, we aspire for the LLM policy tune on the reliable preference data.
So we define a sampling weight for each data.
Given a preference data $\mathcal{D}_{u}^{(i)}$ and each triple has a information gain value $\hat{\mathbb{B}}(\tilde{c}_{j}, \psi|\mathcal{T}_{j}, \mathcal{D}_{u}^{(i)})$, the sampling weight for the current iteration stage $i$ is defined as:
\begin{equation}
\begin{aligned}
\mathcal{P}_j^{(i)} = \frac{(1 - \hat{\mathbb{B}}(\tilde{c}_{j}, \psi|\mathcal{T}_{j}, \mathcal{D}_{u}^{(i)}))\mu}{\sum_{k}(1 - \hat{\mathbb{B}}(\tilde{c}_{k}, \psi|\mathcal{T}_{k}, \mathcal{D}_{u}^{(i)}))\mu},
\label{eql:sampling}
\end{aligned}
\end{equation}
where $\mu>0$ is the hyper-parameter, and $\mathcal{P}_j^{(i)}$ is the probability that the preference triple $(x_j, y_{wj}, y_{lj})$ can be sampled as reliable data, i.e., $\sum_j\mathcal{P}_j^{(i)}=1$.

With the measure of the uncertainty-aware sampling weight, we rewrite the DPO~\footnote{We predominantly focused on DPO in this paper, however, our method can also adapt to PPO in RLHF.} in Eq.~\ref{eql:dpo-loss} to make the LLM capture two kinds of feedback: 1) what responses are better when given a prompt, and 2) what preference triples are better for the LLM to learn preference.
Formally:
\begin{equation}
\begin{aligned}
& \mathcal{L}_{\text{upo}} = -\mathbb{E}_{(x_j, y_{wj}, y_{lj})\sim\mathcal{D}_{u}^{(i)}} \\
& \bigg[\big(1 - \alpha_{j}^{(i)}\big)\log\sigma(\beta h_{\pi_{\theta}^{(i-1)}}^{\pi_{\theta}^{(i)}}) + \alpha_{j}^{(i)}\log\sigma(-\beta h_{\pi_{\theta}^{(i-1)}}^{\pi_{\theta}^{(i)}})\bigg],
\label{eql:upo_loss}
\end{aligned}
\end{equation}
where $h_{\pi_{\theta}^{(i-1)}}^{\pi_{\theta}^{(i)}}$ is the reward margin and defined as:
\begin{equation}
\begin{aligned}
h_{\pi_{\theta}^{(i-1)}}^{\pi_{\theta}^{(i)}} = \log\frac{\pi_{\theta}^{(i)}(y_{wj}|x_j)}{\pi_{\theta}^{(i-1)}(y_{wj}|x_j)} - \log\frac{\pi_{\theta}^{(i)}(y_{lj}|x_j)}{\pi_{\theta}^{(i-1)}(y_{lj}|x_j)}.
\label{eql:upo-reward-difference}
\end{aligned}
\end{equation}
We underscore that $0\leq\alpha_j\leq 1$ is the uncertainty-aware weight for the triple $(x_j, y_{wj}, y_{lj})$ and is used to balance two items in Eq.~\ref{eql:upo_loss}.
In a nutshell, a lower $\alpha_j$ value can encourage the LLM to focus on the given preference data. If the preference data is not reliable according to the uncertainty estimation, we not only expect to reduce the influence of this data but also let the LLM know that the pseudo-labeled preferred response is not suitable and needs to be reversed. 
Thus, we can follow the idea of label smoothing to design the $\alpha_j$ as:
\begin{equation}
\begin{aligned}
\alpha_j = \frac{1}{\mathcal{P}_j + 1}.
\label{eql:upo-alpha}
\end{aligned}
\end{equation}
In addition, to improve the robustness of the iteration preference optimization, we follow~\cite{Pang2024Iterative} to add a negative log-likelihood loss for each preference triple as:
\begin{equation}
\begin{aligned}
\mathcal{L}_{\text{upo+nll}} = \mathcal{L}_{\text{upo}} + \lambda\mathbb{E}_{(x_j, y_{wj}, y_{lj})\sim\mathcal{D}^{(i)}}\frac{\log\pi_{\theta}^{(i)}(y_{wj}|x_j)}{|r_{\phi}^{(i-1)}(x_j, y_{wj})|},
\label{eql:upo-nll}
\end{aligned}
\end{equation}
where $\lambda>0$ is the hyper-parameter.
The whole uncertainty-enhanced self-evolution algorithm is shown in Algorithm~\ref{alg:train}.

\begin{algorithm}[t]
\caption{Uncertainty-Enhanced Self-Evolution}
\label{alg:train}
\begin{small}
\begin{algorithmic}[1]
\REQUIRE LLM SFT model $\pi_{\text{sft}}$, labeled preference data $\mathcal{D}^{(0)}$, prompt set $\mathcal{X}$, iteration total number $I$.
\STATE Training a week LLM policy $\pi_{\theta}^{(0)}$, reward model $r_{\phi}^{(0)}$ and estimator model $f_{\psi}^{(0)}$ over $\mathcal{D}^{(0)}$ based on objectives of Eq.~\ref{eql:dpo-loss}, Eq.~\ref{eql:reward-loss} and Eq.~\ref{eql:estimator-loss}, respectively;
\FOR{iteration state $i\in\{1, 2, \cdots, I\}$}
\STATE Sampling a batch set of prompt $\mathcal{X}_{b}\subset\mathcal{X}$. For each prompt $x_j\in\mathcal{X}_b$, generating at least $N$ responses $\{y_{jk}\}_{k=1}^{N}$;
\STATE Leverage the reward model $r_{\phi}^{(i-1)}$ to assign a score for all responses and pre-screen the permutations to form $\mathcal{D}_{u}^{(i)}$;
\STATE Use the estimator model $f_{\psi}^{(i-1)}$ to perform uncertainty estimation, and obtain probability sampling weight $\mathcal{P}_{j}^{(i)}$ for each prompt in Eq.~\ref{eql:sampling}; 
\STATE Sampling some reliable data $\mathcal{D}_{\text{easy}}^{(i)}$ based on $\mathcal{P}_{j}^{(i)}$, and few unreliable data $\mathcal{D}_{\text{hard}}^{(i)}$ based on $1-\mathcal{P}_{j}^{(i)}$. Thus, we got the final preference data as $\mathcal{D}^{(i)}=\mathcal{D}_{\text{easy}}^{(i)}\cup\mathcal{D}_{\text{hard}}^{(i)}$;
\STATE Sequentially update the LLM Policy, reward, and estimator model by Eq.~\ref{eql:upo-nll}, Eq.~\ref{eql:reward-loss}, and Eq.~\ref{eql:estimator-loss}, respectively.
\ENDFOR
\RETURN The LLM policy model $\pi_{\theta}^{(I)}$.
\end{algorithmic}
\end{small}
\end{algorithm}

\section{Experiments}

In this section, we choose universal NLP and mathematics reasoning tasks to evaluate the effectiveness of the UPO framework.

\subsection{Universal NLP Tasks}
Following the practice in previous works, we validate the performance of LLM policy trained through the UPO framework over AlpacaEval 2.0~\cite{Dubois2024Length} and MT-Bench~\cite{Zheng2023Judging}.
The benchmark of AlpacaEval 2.0 consists of 805 instructions and can be used to approximately head-to-head test the length-controlled (LC) weighted win rate of preference annotated by GPT-4.
MT-Bench aims to evaluate the capability (scoring from 0 to 10) of the LLM policy to solve multiple basic problems such as writing, roleplay, reasoning, math, coding, extraction, stem, and humanities.

For the implementation setups, we choose \texttt{zephyr-7b-sft-full} (default as Zephyr-7B) as the backbone, which has been further instruction-tuned over UltraChat200K dataset
from Mistral-7B~\cite{Jiang2023Mistral}.
The labeled preference data we used is UltraFeedback~\cite{Cui2023UltraFeedback}, which consists of 61K prompts post-processed by~\citet{Tunstall2023Zephyr}
.
We also select UltraChat200K as the prompt set.
We repeatedly train three models (i.e., LLM policy, reward, and estimator) for three iterations. 
For the baselines, we choose SFT and DPO trained from Zephyr-7B to make a comparison.
In addition, we also collect all cleaned preference data from the initial stage and three iterations and use DPO to train a model as UPO-Merge.
More details of these benchmarks and hyper-parameters of each training iteration are listed in Appendix~\ref{sec:implementation}.

\begin{table}[t]
\centering
\resizebox{\linewidth}{!}{
\begin{small}
\begin{tabular}{l | c |cc}
\toprule
\bf Models &\bf  Align &\bf AlpacaEval 2.0  &\bf MT-bench \\
\midrule
Mistral-7B & no & 0.17 & 3.25 \\
Alpaca-7B & no & 5.88 & 5.81 \\
\midrule
Zephyr-7B-SFT  & no & 5.84 & 6.18 \\
Zephyr-7B-DPO & yes & 9.12& 6.79 \\
Zephyr-7B-UPO & yes & \textbf{13.04} & \textbf{7.02} \\
Zephyr-7B-UPO-Merge & yes & 12.04 & 6.85 \\
\bottomrule
\end{tabular}
\end{small}
}
\caption{Main results derived from GPT-4 auto evaluation on AlpacaEval 2.0 (LC weighted win rate \% compared with reference of GPT-4) and MT-Bench (absolute score).}
\label{tab:main-results-nlp}
\end{table}

\paragraph{Main Results}
As shown in Table~\ref{tab:main-results-nlp}, the results of AlpacaEval 2.0 denote the win rate compared to the reference generated by GPT-4, and we can see that the LLM policy of Zephyr-UPO after three iterations achieves the best win rate against GPT-4 and improves by 7.20\% and 3.92\% over SFT and DPO, respectively.
To further investigate the performance at each iteration compared to the baseline, we use GPT-4 to annotate the preference for each iteration and present in Table~\ref{tab:main-results-head-alpaca}. The results suggest that the best performance can be achieved at the second iteration and improved by over 20\%.
It is noteworthy that the performance improvement does not rely on increasing response length, which indicates that our method can empower the output quality of LLM instead of outputting long text. 
For the benchmark of MT-Bench, we also use GPT-4 to annotate the average score of eight aspects and the results in Table~\ref{tab:main-results-nlp} show that our method can obtain the highest score and improve the LLM policy from 6.79\% to 7.02\%.

In addition, by comparing the performance of UPO-Merge with DPO and UPO, we can obtain the following suggestions:
1) the result of UPO-Merge is lower than UPO, which means that iterative evolution is more effective than single turn even though post-train with the same number of preference data, and
2) expending the preference data by self-generation manner can substantially enhance the LLM policy on universal NLP ability.

\begin{table}[t]
\centering
\resizebox{\linewidth}{!}{
\begin{small}
\begin{tabular}{l | c |cc}
\toprule
\bf Models &\bf  Align &\bf Win Rate  &\bf Avg. Length \\
\midrule
Zephyr-7B-SFT  & no & 50.00 & 1014 \\
Zephyr-7B-DPO & yes & 66.40 & 1298 \\
Zephyr-7B-UPO-Iter1 & yes & 69.94 & 967 \\
Zephyr-7B-UPO-Iter2 & yes & \textbf{71.53} & \textbf{1148} \\
Zephyr-7B-UPO-Iter3 & yes & 70.21 & 1162 \\
Zephyr-7B-UPO-Merge & yes & 70.39 & 1200 \\
\bottomrule
\end{tabular}
\end{small}
}
\caption{Main results derived from  GPT-4 auto evaluation (LC weighted win rate \%) of different iterations model from UPO over AlpacaEval 2.0 head-to-head comparison with responses of Zephyr-7B-SFT.}
\label{tab:main-results-head-alpaca}
\end{table}

\begin{table}[t]
\centering
\resizebox{\linewidth}{!}{
\begin{small}
\begin{tabular}{l | c |cc}
\toprule
\bf Models &\bf  Align &\bf GSM8K  &\bf MATH \\
\midrule
GPT-4o-0513 & yes &  95.8 & 76.6 \\
Claude-3-Opus & yes &  95.0 & 60.1 \\
Gemini-1.5-Pro (May) & yes &  90.8 & 67.7 \\
Qwen2-7B-Instruct & yes &  82.3 & 49.6 \\
\midrule
Qwen2-7B-SFT$^\dagger$  & no &  88.2 & 54.8 \\
Qwen2-7B-DPO & yes & 88.3 & 55.0 \\
Qwen2-7B-StepDPO$^\dagger$ & yes &  88.5 & 55.8 \\
\midrule
Qwen2-7B-UPO-Iter1 & yes & 88.5 & 55.4 \\
Qwen2-7B-UPO-Iter2 & yes & 88.6 & 55.7 \\
Qwen2-7B-UPO-Iter3 & yes & 88.4 & 55.6 \\
Qwen2-7B-UPO-Merge & yes & 88.4 & 55.6 \\
\midrule
Qwen2-7B-StepUPO-Iter1 & yes & 88.8 & 56.0 \\
Qwen2-7B-StepUPO-Iter2 & yes & \textbf{88.9} & \textbf{56.3} \\
Qwen2-7B-StepUPO-Iter3 & yes & 88.8 & 56.1 \\
Qwen2-7B-StepUPO-Merge & yes & 88.8 & 56.2 \\
\bottomrule
\end{tabular}
\end{small}
}
\caption{Main results (accuracy \%) on GSM8K and MATH benchmarks. $^\dagger$ is trained by~\cite{Lai2024Step}.}
\label{tab:main-results-math}
\end{table}

\subsection{Mathematics Reasoning}
Apart from the universal generation, we also choose two widely-used GSM8K~\cite{cobbe2021training} and MATH~\cite{Hendrycks2021Measuring} to show the versatility of UPO on complex reasoning benchmarks.
GSM8K consists of 8.5K high-quality linguistically diverse grade school math word problems and requires the LLM policy to multi-step reasoning capability, while MATH aims at featuring challenging competition math problems.

For the implementation, we choose MathInstruct~\cite{Yue2024MAmmoTH} as the prompt set which focuses on the hybrid use of chain-of-thought (CoT) and program-of-thought (PoT) rationales. It contains 262K prompts that are compiled from 13 math rationale datasets. We remove GSM8K and MATH from it to prevent the data leak problem.
We follow~\citet{Lai2024Step} to use the technique of StepDPO to tune the LLM policy and the well-constructed fine-grained feedback data is Math-Step-DPO-10K
which involves 10.8K prompts with both coarse-grained and fine-grained annotation towards the answers.
We select \texttt{Qwen2-7B-SFT} and \texttt{Qwen2-7B-SFT-Step-DPO} as our basic backbones $\pi_{\text{sft}}$ and the initial LLM policy $\pi_{\theta}^{(0)}$, respectively.
The model trained based on our framework with DPO and StepDPO paradigms are respectively named as UPO and StepUPO.
During the iteration, we do not filter the noisy data by directly matching the ground truth of each reasoning step or the final answer. In other words, we only leverage the uncertainty estimator to verify the reliable of each reasoning step, aiming to simulate the real scenario that solves the unseen question.
More details of these benchmarks and training setups are shown in Appendix~\ref{sec:implementation-2}.

\begin{table}[t]
\centering
\resizebox{\linewidth}{!}{
\begin{small}
\begin{tabular}{l | cc | cc}
\toprule
\bf \multirow{2}{*}{\textbf{Models}} &\bf AlpacaEval 2.0  &\bf MT-bench  & \bf GSM8K & \bf MATH \\
& \multicolumn{2}{c|}{Zephyr-7B} & \multicolumn{2}{c}{Qwen2-7B} \\
\midrule
SFT   & 5.84 & 6.18 & 88.2 & 54.8 \\
DPO / StepDPO & 9.12& 6.79 & 88.5 & 55.8 \\
\midrule
UPO / StepUPO  & \textbf{13.04} & \textbf{7.02} & \textbf{88.9} & \textbf{56.3} \\
w/o. Rule  & 13.01  & 7.01 & 88.8 & 56.1 \\
w/o. Estimator & 10.84 & 6.52 & 87.1 & 54.7 \\
w/o. Weight $\alpha$ & 12.70 & 6.94 & 88.0 & 55.8 \\
w/o. NLL loss & 12.39 & 6.92  & 87.9 & 55.7 \\

\bottomrule
\end{tabular}
\end{small}
}
\caption{Ablation study at the first iteration over AlpacaEval 2.0 (LC weighted win rate \% compared with GPT-4), MT-Bench (absolute score), GSM8K (accuracy \%) and MATH (accuracy \%).}
\label{tab:ablation}
\end{table}

\begin{figure}[t]
\centering
\begin{tabular}{cc}
\begin{minipage}[t]{0.5\linewidth}
    \includegraphics[width = \linewidth]{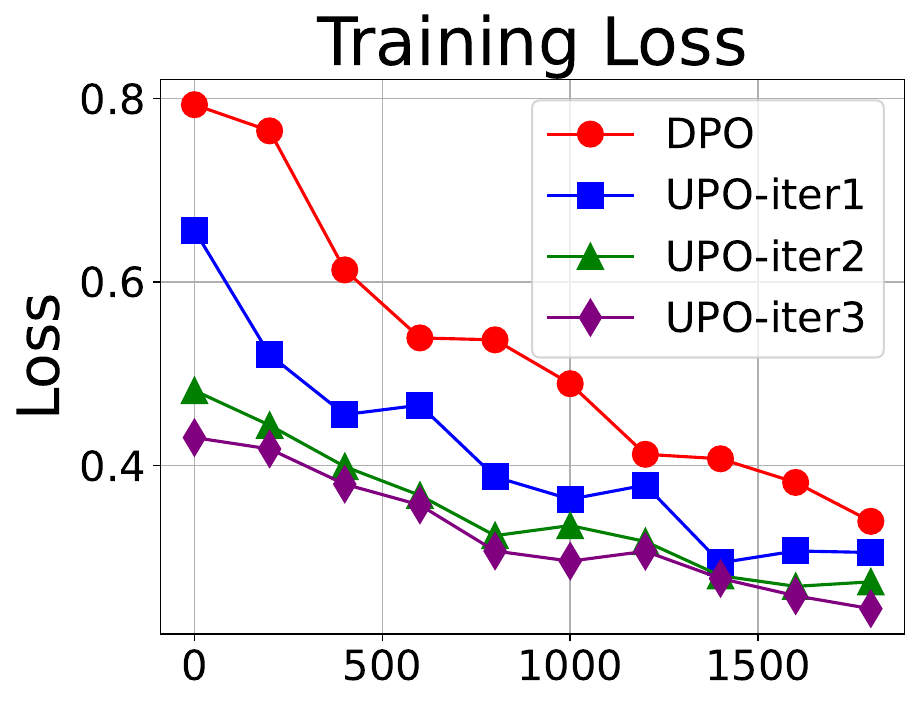}
\end{minipage}
\begin{minipage}[t]{0.5\linewidth}
    \includegraphics[width = \linewidth]{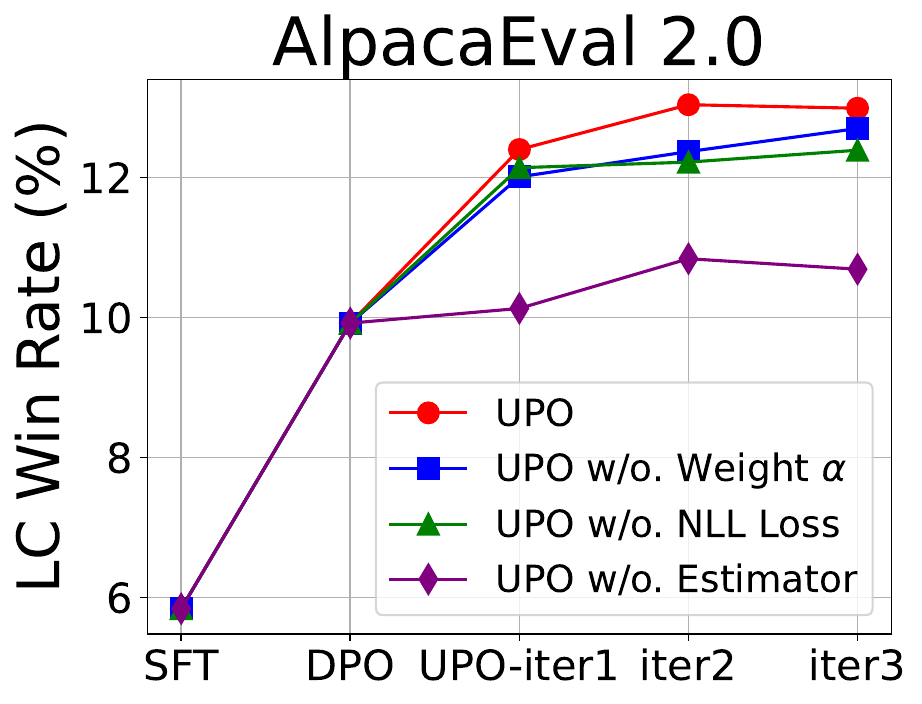}
\end{minipage}
\end{tabular}
\caption{The curve of training loss and LC win rate (\%) on AlpacaEval 2.0 at each iteration.}
\label{fig:convergence-loss-alpacaeval}
\end{figure}

\begin{figure}[t]
\centering
\includegraphics[width=\linewidth]{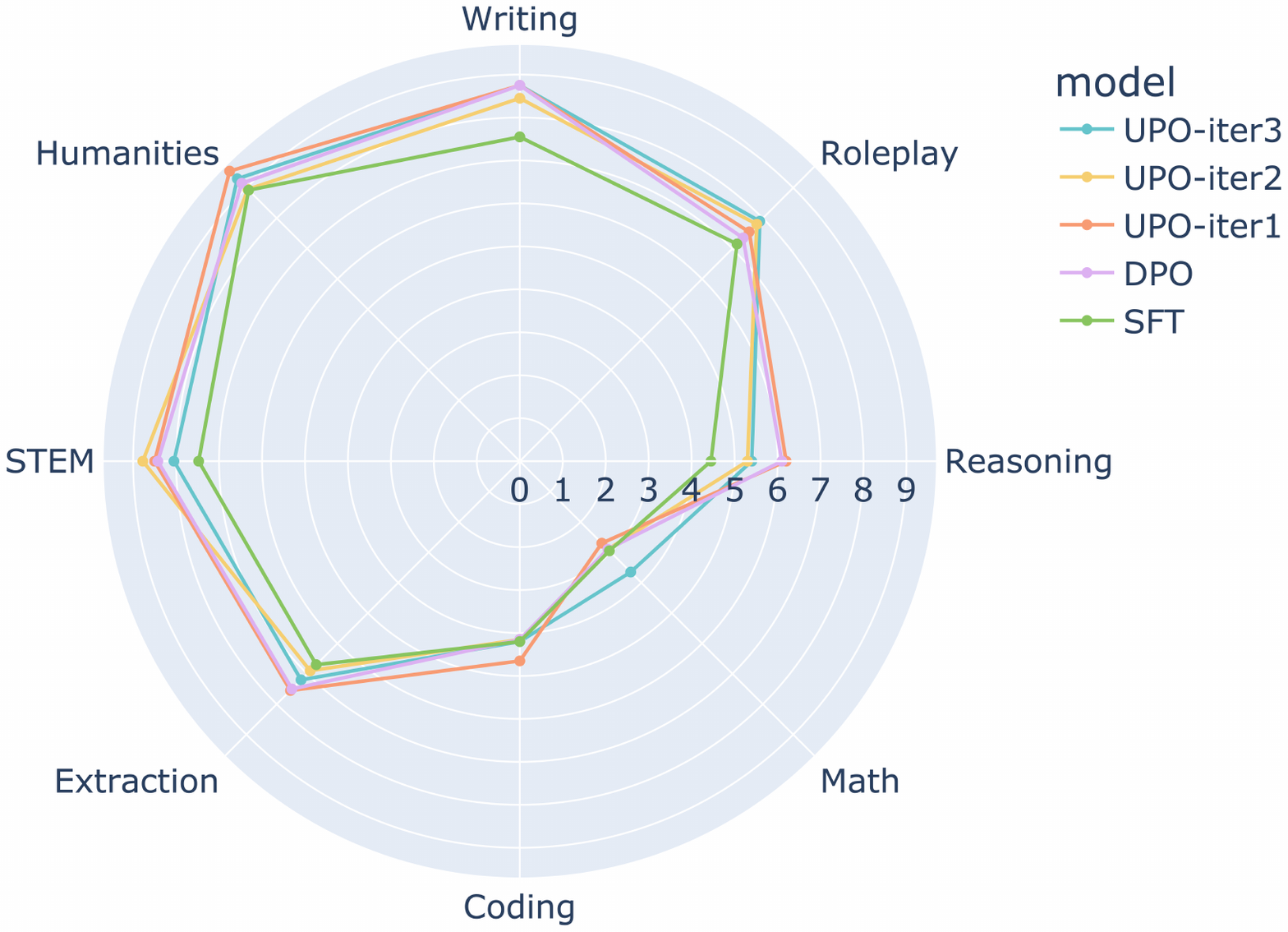}
\caption{Performance of different iterations of UPO compared with SFT and DPO over MT-Bench.}
\label{fig:mt-bench-aspect}
\end{figure}

\paragraph{Main Results}
The results are listed in Table~\ref{tab:main-results-math} and we can obtain the following suggestions:
1) The LLM policy post-trained by DPO makes a marginal improvement, increasing from 88.2\% and 54.8\% to 88.3\% and 55.0\%, respectively.
Yet, the improvement of StepDPO can achieve an obvious gain compared with the SFT model, indicating that LLM policy self-evolution can be better conducted with fine-grained feedback.
2) For each iteration, UPO and StepUPO can consistently achieve substantial improvements on GSM8K and MATH, respectively resulting in 88.9\% and 56.3\% accuracy metrics.
3) The result of UPO-Merge and StepUPO-Merge is similar to the performance at the third iteration, which conflicts with the findings in universal NLP tasks. We analyze that the task of mathematics reasoning highly relies on the cleaned preference data, yet the preference data after uncertainty estimation may still contain noisy fine-grained feedback and affect the performance inevitably.

\begin{figure*}[t]
\centering
\begin{tabular}{cc}
\begin{minipage}[t]{0.33\linewidth}
    \includegraphics[width = .9\linewidth]{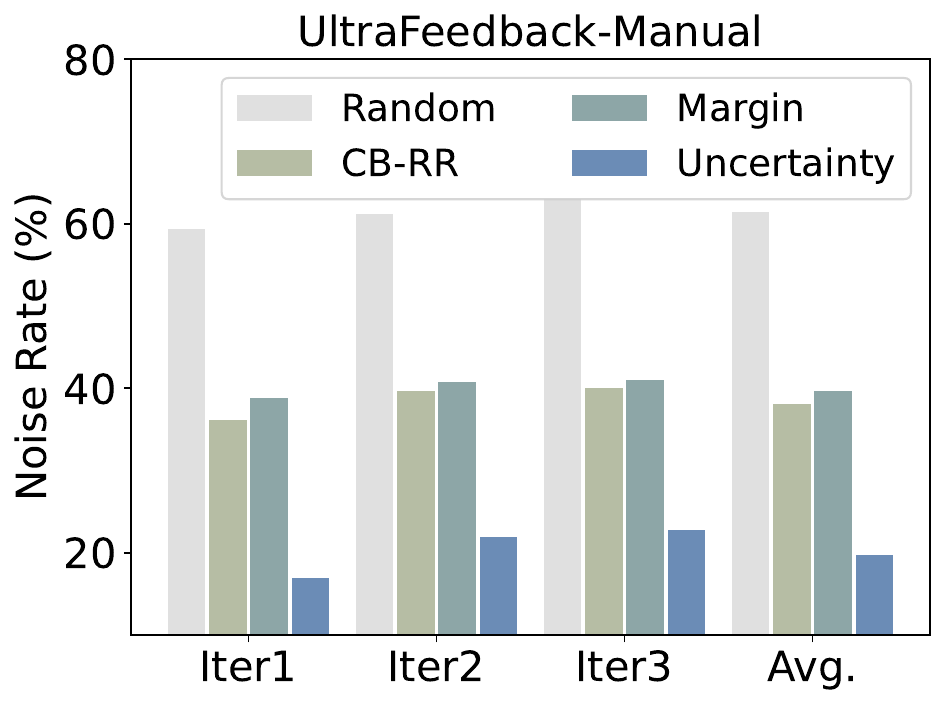}
\end{minipage}
\begin{minipage}[t]{0.33\linewidth}
    \includegraphics[width = .9\linewidth]{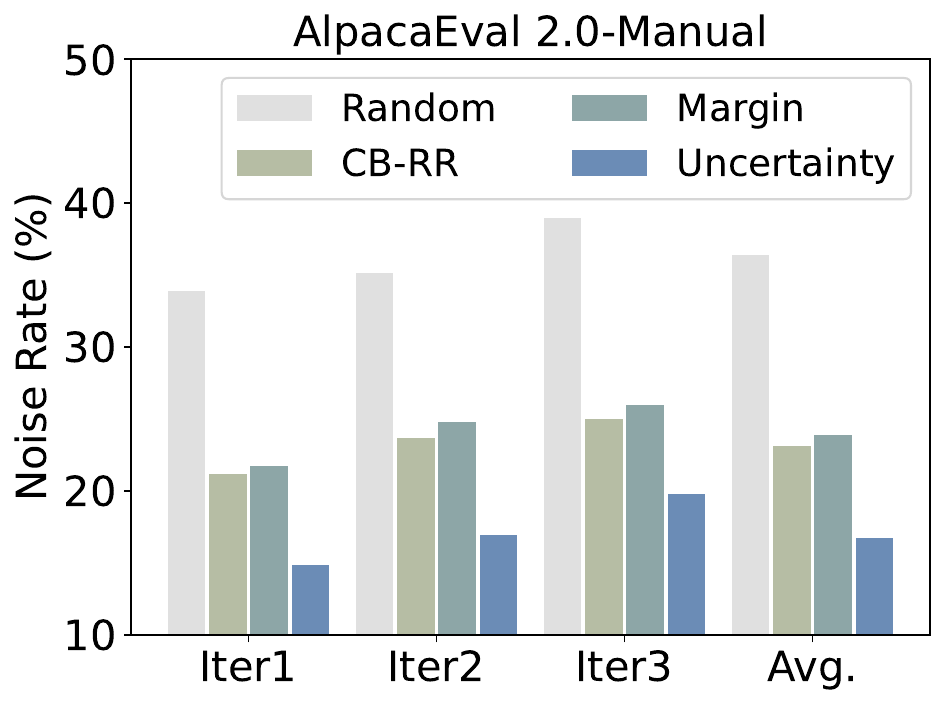}
\end{minipage}
\begin{minipage}[t]{0.33\linewidth}
    \includegraphics[width = .9\linewidth]{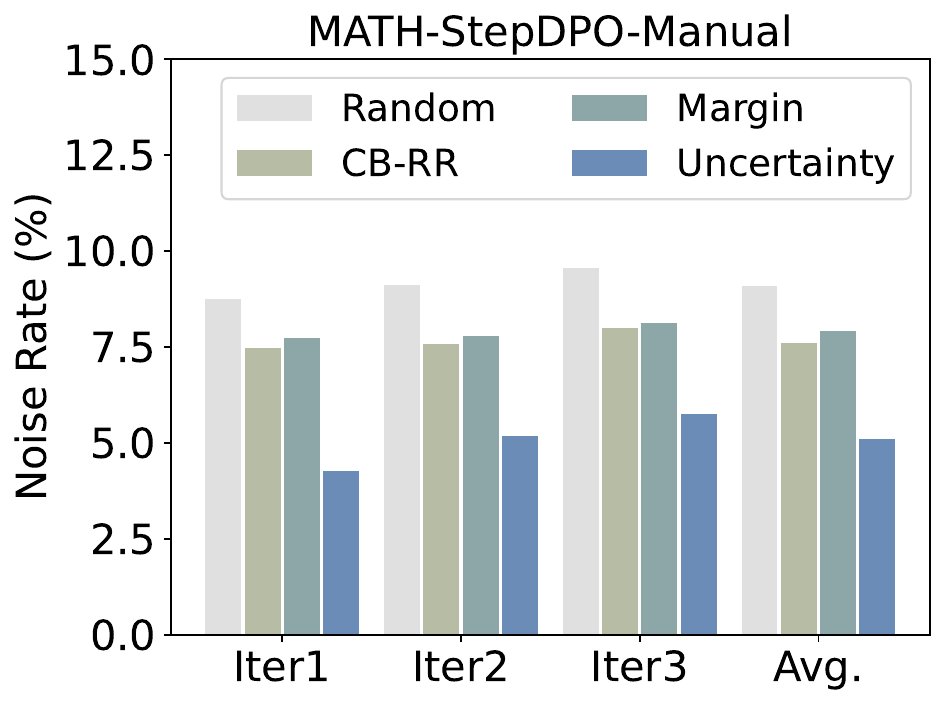}
\end{minipage}
\end{tabular}
\caption{Noise rate (\%) of different sampling strategies over multiple manual evaluation sets.}
\label{fig:noise-rate}
\end{figure*}

\section{Further Analysis}
\subsection{Ablation Study}
To investigate the impact of different techniques used in UPO, we conduct the ablation study on all benchmarks to see the performance of different variants.
Specifically, for benchmarks of AlpacaEval 2.0 and MT-Bench, we choose DPO as the main baseline and optimization paradigm, while the StepDPO paradigm will be used in GSM8K and MATH. We conduct the experiments at the first iteration.
For the variants, w/o. Rule means directly choosing all permutations without any pre-screen processing.
w/o.  Estimator denotes that do not use uncertainty estimation and choose all generated preference data to train the LLM policy, which is the same as vanilla iterative preference optimization proposed by~\cite{Pang2024Iterative}.
w/o. Weight $\alpha$ represents only training the LLM policy on DPO or StepDPO without smoothing (i.e., $\alpha=0$).
w/o. NLL loss means removing the NLL loss by setting $\lambda=0$.
Results demonstrated in Table~\ref{tab:ablation} show that the performance will drop if the framework module is removed.
Moreover, the use of robust techniques (i.e., uncertainty-enhanced weighting and the NLL loss) consistently contributes to the robustness improvement when training on pseudo preference data.

\subsection{Effectiveness of Uncertainty-Enhanced Self-evolution}
We also explore how the Uncertainty-Enhanced Self-evolution algorithm empowers the LLM policy in the iteration preference optimization procedure.
To ask this question, we choose the benchmarks of AlpacaEval 2.0 and MT-Bench to make a deep-seek.
We first draw a training loss curve at the initial stage (DPO training) and each iteration in UPO when preference optimizing on UltraFeedback and newly generated preference data sampled from UltraChat200K.
The curve presented in Figure~\ref{fig:convergence-loss-alpacaeval} (left) demonstrates that iterative procedure advances the convergence which may contribute to the high performance.

To see the performance changes in different training stages, we also draw a curve to show the win rate increasing in Figure~\ref{fig:convergence-loss-alpacaeval} (right) with multiple variants.
The result suggests that UPO can substantially outperform vanilla preference optimization (e.g., DPO) in all iteration stages.
It is worth noting that variant UPO w/o. Estimator has a bit of improvement compared to the DPO,  indicating that many noisy pseudo-preference examples are used in the next iteration and make the iteration training useless. 
This finding reflects that the noisy reduction and robustness consideration in iteration preference optimization is significantly necessary.

\subsection{Capability Across Different Aspects in MT-Bench}
To show the performance of the LLM policy tuned by the UPO framework, we perform task-wise deep analysis on MT-Bench and show the capability of eight aspects in Figure~\ref{fig:mt-bench-aspect}, including writing, roleplay, reasoning, math, coding, extracting, STEM, and humanities.
Results show that UPO consistently enhances the generation of LLM policy on different aspects of basic problems. 
Notably, UPO can also realize an obvious improvement in complex tasks, such as reasoning, math, and coding.

\subsection{Noisy Data Study}
We end this section by investigating how the UPO framework realizes denoising during iteration preference optimization.
We respectively sample 200 preference data from the validation set of UltraFeedback, AlpacaEval 2.0, and MATH-Step-DPO-10K to manually construct the evaluation set.
In particular, for preference data from UltraFeedback and MATH-Step-DPO-10K, we directly use the label (which response is better) as the ground truth. For AlpacaEval 2.0, we use the reference generated from GPT-4 as the preferred response, while the dispreferred response is created by the SFT model.
At each iteration, we present four different reliable data sampling strategies to select preference data to train the LLM policy after the rewarding process.
1) ``Random'' denotes randomly selecting from pseudo preference data;
2) ``CB-RR'' means \textbf{C}hosen response with \textbf{B}est reward and \textbf{R}ejected response with \textbf{R}andom select from the rest lower reward, which is a similar strategy to UltraFeedback.
3) ``Margin'' denotes choosing only one preference data whose reward margin between chosen and rejected is the largest.
4) ``Uncertainty'' is our proposed method that uses the certainty weight to perform sampling.

Results demonstrated in Figure~\ref{fig:noise-rate} indicate that considering the reward of the chosen response or reward margin is certainly effective to denoising, which has also been proven in some previous work~\cite{Pang2024Iterative}.
In addition, the results also showcase that leveraging uncertainty estimation can better reduce the noise rate by more than 20\%, 10\%, and 3\%, respectively, indicating the effectiveness of UPO.






\section{Related Works}
\subsection{Preference Optimization of LLMs}

Large language models (LLMs), after undergoing extensive pre-training, may generate fabricated facts, biased content, or harmful text. To align these models with human values, fine-tuning language models to adhere to human preferences is an effective solution. Reinforcement Learning from Human Feedback (RLHF) ~\cite{stiennon2020learning,ziegler2019fine} has emerged as a groundbreaking technique for aligning LLMs. By training a reward model on human feedback data and using Proximal Policy Optimization (PPO)~\cite{schulman2017proximal} to obtain the policy model for language generation, this approach has led to the development of powerful models such as GPT-4~\cite{achiam2023gpt}, Llama3~\cite{dubey2024llama}, and Gemini~\cite{team2023gemini}.
Other methodologies such as DPO~\cite{rafailov2024direct} and RRHF~\cite{yuan2023rrhf}, optimize language models directly on human feedback datasets. Nevertheless, to further improve performance, it becomes essential to conduct sampling using the model itself, necessitating the incorporation of an auxiliary reward model (RM) ~\cite{liu2023statistical,song2024preference,zhou2023beyond,dong2023raft,touvron2023llama}.

\subsection{Iterative Preference Optimization}
The optimization of preference datasets and preference models plays a significant role in the alignment of LLMs.
Some works~\cite{dong2023steerlm, wang2024helpsteer, rame2024rewarded} employ fine-grained reward objectives and iteratively fine-tune large models for alignment.
For example, IRPO~\cite{Pang2024Iterative}, utilizes iterative DPO for optimization.\cite{Yuan2024Self} directly explores a novel Self-Rewarding method for LLMs, which achieve self-improvement by generating their rewards during training.
~\cite{fisch2024robust} proposes a reward model distillation algorithm to address the effectiveness and robustness in preference optimization.
Similar to these works, we also focus on how to iteratively enhance the effectiveness of preferences and address the noise in the preference predictions by the reward model, aiming to improve the overall robustness of the alignment process.




\section{Conclusion}

We propose an uncertainty-enhanced preference optimization framework to further boost the abilities of the self-evolution of LLMs.
We develop an estimator model and let it cooperate with the reward model to provide high-quality preference data at each iteration stage.
To reach this goal, we leverage the MC Dropout technique in BNN to perform uncertainty estimation, eliminating the potentially noisy data derived from the weak LLM policy.
In addition, we also propose an uncertainty-enhanced self-evolution algorithm to improve the robustness of LLM when repeatedly updating parameters via DPO.
We conduct extensive experiments on multiple universal NLP and mathematics reasoning tasks and the results indicate the effectiveness of our method.
In the future, we aim to further improve the overall performance and adapt the framework to PPO and other LLMs.

\bibliography{custom}

\appendix

\section{Prompt Template for Estimator}
\label{sec:app-prompt}

\begin{figure}[h]
\centering
\includegraphics[width=\linewidth]{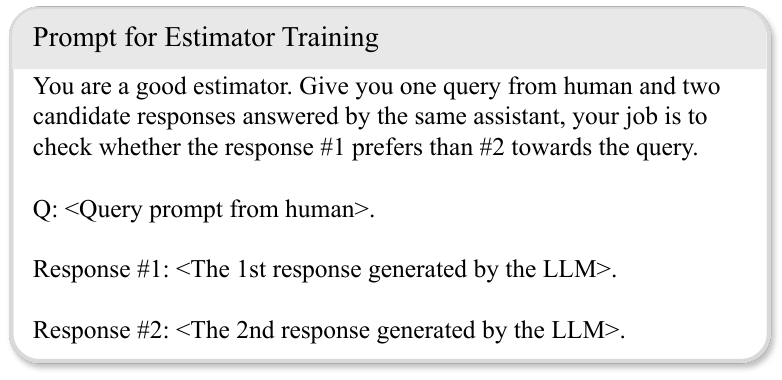}
\caption{Prompt format of the estimator input.}
\label{fig:estimator-prompt}
\end{figure}

The prompt template used for the estimator model is shown in Figure~\ref{fig:estimator-prompt}.
During the training stage, we leverage the objective of ``AutoModelForClassification'' implemented by the Transformers toolkit. We use the representation of the last token \texttt{[EOS]} to make classification.
It is worth noting that, we do not transform the objective of the estimator to an instruction-following task because it has a different distribution in the label space (binary space v.s. vocabulary space).

\section{Implementation Setups of Universal NLP Tasks}
\label{sec:implementation}
We provide the implementation setup details for the experiment of universal NLP tasks.

\subsection{Initial Stage}
In the initial stage, we use processed UltraFeedback~\footnote{\url{https://huggingface.co/datasets/HuggingFaceH4/ultrafeedback_binarized}.} as the seed preference data, comprising 61k prompts with annotated preference pairs by AI feedback. 
This data will be used to train a week LLM policy $\pi_{\theta}^{(0)}$, reward model $r_{\phi}^{(0)}$ and estimator model $f_{\psi}^{(0)}$.
The backbone we used is~\texttt{Zephyr-7b-sft-full}.

\begin{table}[t]
\centering
\resizebox{\linewidth}{!}{
\begin{small}
\begin{tabular}{l | c | c | c}
\toprule
\bf Param. &\bf  Policy &\bf Reward &\bf Estimator \\
\midrule
backbone & zephyr-7b & zephyr-7b & zephyr-7b \\
batch & 8/gpu*8gpus & 8/gpu*8gpus & 8/gpu*8gpus \\
epoch & 3 & 3 & 2 \\
lr  & 5e-6 & 1e-4 & 2e-5 \\
scheduler  & linear & linear & linear \\
warm rate  & 0.1 & 0.1 & 0.1 \\
use bf16 & yes & yes & yes \\
use Lora  & yes & yes & yes \\
Lora modules  & all & all & all \\
length & 1024 & 1024 & 2048 \\
$\beta$ & 0.1 & - & - \\
\bottomrule
\end{tabular}
\end{small}
}
\caption{The hyper-parameters used in the initial stage over universal NLP tasks.}
\label{tab:hyper-param-initial-nlp}
\end{table}

\begin{table}[t]
\centering
\begin{small}
\begin{tabular}{l | c | c | c}
\toprule
\bf Param. &\bf  Iter1 &\bf Iter2 &\bf Iter3 \\
\midrule
batch & 8/gpu*8gpus & 8/gpu*8gpus & 8/gpu*8gpus \\
epoch & 3 & 3 & 3 \\
lr  & 5e-6 & 4e-6 & 3e-6 \\
scheduler  & linear & linear & linear \\
warm rate  & 0.1 & 0.1 & 0.1 \\
use bf16 & yes & yes & yes \\
use Lora  & yes & yes & yes \\
Lora modules  & all & all & all \\
length & 1024 & 1024 & 1024 \\
$\beta$ & 0.1 & 0.1 & 0.1 \\
$\lambda$ & 1.0 & 1.0 & 1.0 \\
\bottomrule
\end{tabular}
\end{small}
\caption{The hyper-parameters of LLM policy used in the different iteration stages over universal NLP tasks.}
\label{tab:hyper-param-iteration-nlp}
\end{table}

\begin{figure*}[t]
\centering
\begin{tabular}{cc}
\begin{minipage}[t]{0.5\linewidth}
    \includegraphics[width = \linewidth]{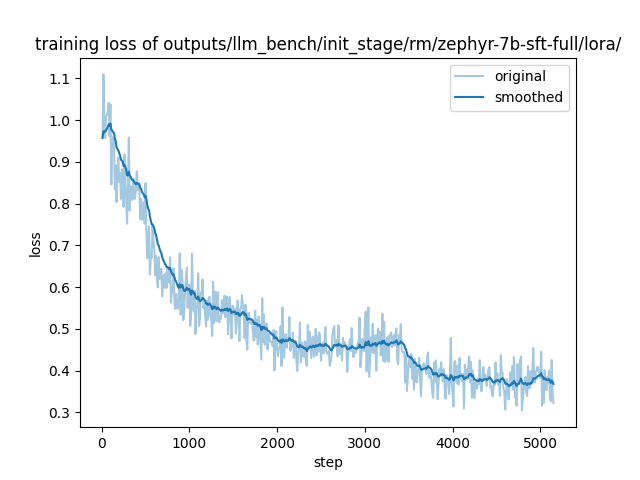}
\end{minipage}
\begin{minipage}[t]{0.5\linewidth}
    \includegraphics[width = \linewidth]{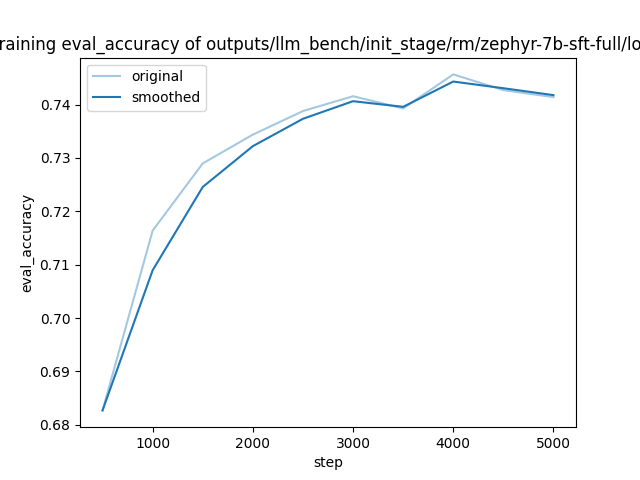}
\end{minipage}
\end{tabular}
\begin{tabular}{cc}
\begin{minipage}[t]{0.5\linewidth}
    \includegraphics[width = \linewidth]{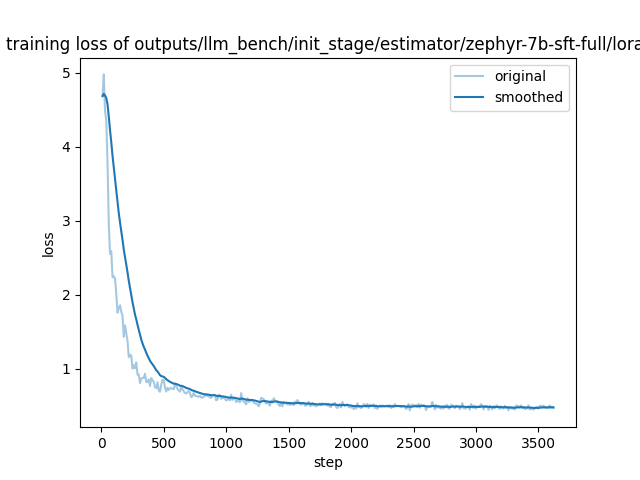}
\end{minipage}
\begin{minipage}[t]{0.5\linewidth}
    \includegraphics[width = \linewidth]{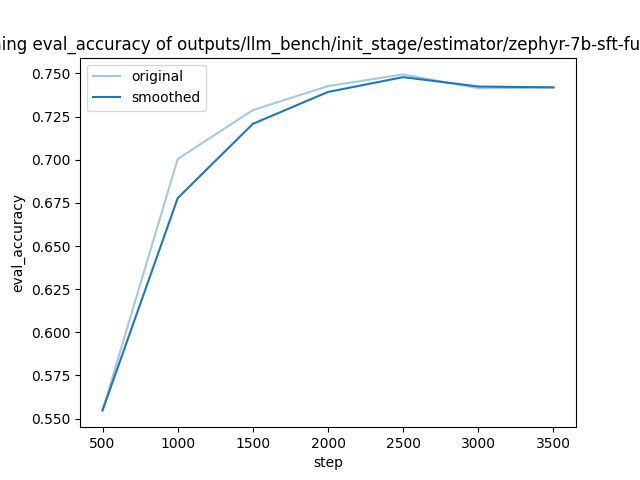}
\end{minipage}
\end{tabular}
\caption{The curves of training loss and evaluation accuracy (\%) for the reward model (the first row) and estimator model (the last row) on UltraFeedback preference data at the initial stage.}
\label{fig:nlp-rm-es-curve}
\end{figure*}

To train a week LLM policy, we directly use the vanilla DPO algorithm~\cite{Rafailov2023Direct}, and the backbone is borrowed from zephyr-7b. We also use this backbone to train a reward model and an estimator model. The training parameters are shown in Table~\ref{tab:hyper-param-initial-nlp}.
We draw the curve of training loss and evaluation accuracy for the reward model and estimator model in Figure~\ref{fig:nlp-rm-es-curve}.


\subsection{Iteration Stage}
At each iteration, we randomly sample 10k prompts from UltraFeedback and 25k prompts from the set of UltraChat200K~\footnote{\url{https://huggingface.co/datasets/HuggingFaceH4/ultrachat_200k}.}, with final total 35k prompts.
We then prompt the LLM policy at the last iteration to generate at least 4 outputs for each prompt, and then assign the reward score by the reward model.
The temperature and topp values we used are 0.8 and 0.9, respectively.

After that, we can obtain multiple permutations and feed them into the estimator model to perform uncertainty estimation.
To be specific, the MC Dropout rate we used is 0.1 for Lora, the inference times $T=10$. We sample 50\% data from the pseudo preference data after the estimation as the easy set, and randomly sampled 40\% data from the original seed preference data. We can get about 35k preference data at each iteration.

In order to make the overall framework training efficient, we only update the parameters of LLM policy on the newly constructed preference data. The hyper-parameters of the LLM policy at each iteration are shown in Table~\ref{tab:hyper-param-iteration-nlp}.

All experiments at initial stage and iteration stage are based on 8 GPUs NVIDIA A100 (80G). The whole framework with 3 iteration stages takes 2.5 days.

\begin{table}[t]
\centering
\resizebox{\linewidth}{!}{
\begin{small}
\begin{tabular}{l | c | c | c}
\toprule
\bf Param. &\bf Policy &\bf Reward &\bf Estimator \\
\midrule
backbone & qwen2-7b & qwen2-7b & qwen2-7b \\
batch & 8/gpu*4gpus & 8/gpu*4gpus & 8/gpu*4gpus \\
epoch & 3 & 3 & 3 \\
lr  & 5e-6 & 1e-4 & 2e-5 \\
scheduler  & linear & linear & linear \\
warm rate  & 0.1 & 0.1 & 0.1 \\
use bf16 & yes & yes & yes \\
use lora  & no & no & no \\
unfreeze layers  & upper 8 & upper 8 & upper 8 \\
length & 1024 & 1024 & 2048 \\
$\beta$ & 0.1 & - & - \\
\bottomrule
\end{tabular}
\end{small}
}
\caption{The hyper-parameters used in the initial stage over universal Mathematics tasks.}
\label{tab:hyper-param-initial-math}
\end{table}

\section{Implementation Setups of Mathematics Reasoning Tasks}
\label{sec:implementation-2}
Next, we provide the implementation details of the mathematics reasoning tasks.
We use open-source training data of ``math-step-dpo-10k''~\footnote{\url{https://hf-mirror.com/datasets/xinlai/Math-Step-DPO-10K}.}  released by~\citet{Lai2024Step}, which consists of about 10k fine-grained preference data. For each example, the preferred response (chosen) and dispreferred response (rejected) share the same prefix reasoning steps which are correct toward the prompt, the negative step can be sampled by the self-consistency method.

\subsection{Initial Stage}
In the first stage, we utilize preference data from math-step-dpo-10k to train both reward and estimator models. 
The backbone we used is~\texttt{Qwen2-7B}.
As noted earlier, we present a StepUPO variant, which aims to expand the StepDPO iteratively. The primary distinction lies in the structure of the preference data. 
Specifically, the data utilized for StepDPO must consist of step-by-step fine-grained preference feedback, while the data for DPO is based on sentence-wise preference feedback. We have observed that the data from math-step-dpo-10k also includes sentence-by-sentence feedback. Therefore, we can employ it to train the original DPO-based LLM policy, denoted as Qwen2-7B-DPO. 
As for the StepDPO-based LLM policy, we directly utilize the trained~\texttt{Qwen2-7B-SFT-Step-DPO} as the LLM policy.

\begin{figure*}
\centering
\begin{tabular}{cc}
\begin{minipage}[t]{0.5\linewidth}
    \includegraphics[width = \linewidth]{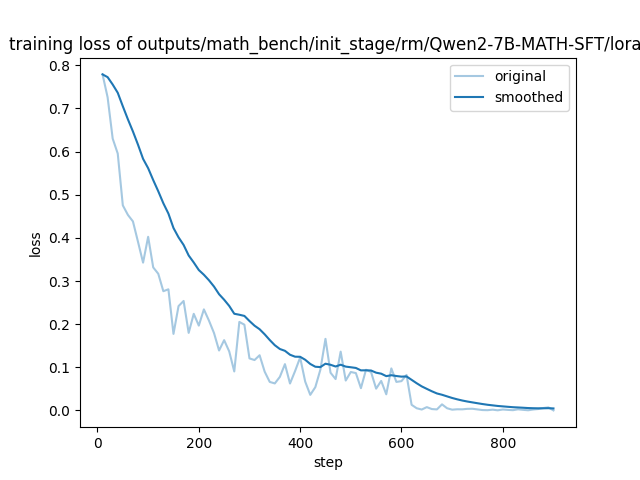}
\end{minipage}
\begin{minipage}[t]{0.5\linewidth}
    \includegraphics[width = \linewidth]{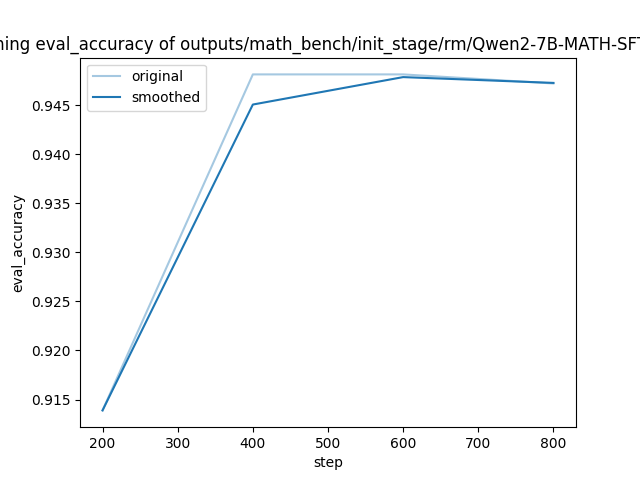}
\end{minipage}
\end{tabular}
\begin{tabular}{cc}
\begin{minipage}[t]{0.5\linewidth}
    \includegraphics[width = \linewidth]{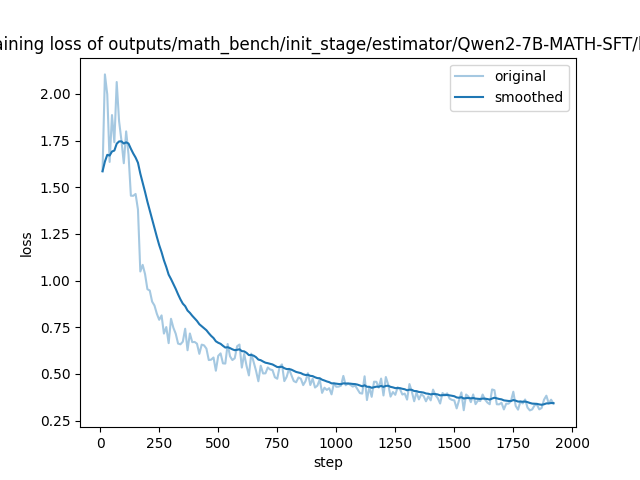}
\end{minipage}
\begin{minipage}[t]{0.5\linewidth}
    \includegraphics[width = \linewidth]{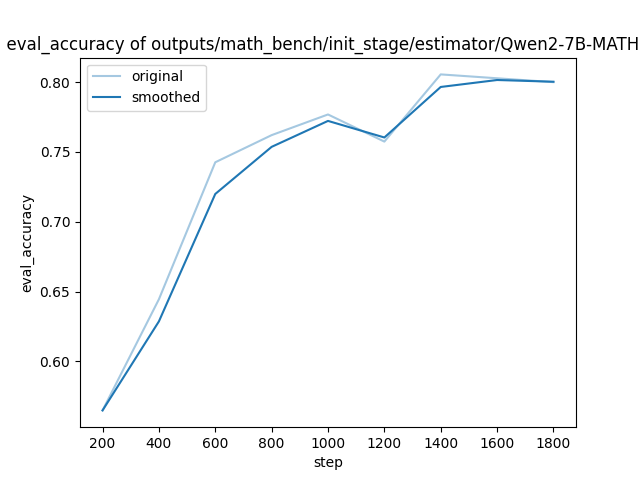}
\end{minipage}
\end{tabular}
\caption{The curves of training loss and evaluation accuracy (\%) for the reward model (the first row) and estimator model (the last row) on Math-Step-DPO-10K preference data at the initial stage.}
\label{fig:math-rm-es-curve}
\end{figure*}

The details of training setups are shown in Table~\ref{tab:hyper-param-initial-math}.
The curve of training loss and evaluation accuracy are shown in Figure~\ref{fig:math-rm-es-curve}.
We do not equip the Lora module in the backbone because the vocabulary set is too large to support vLLM acceleration. To simulate the parameter-efficient learning, we unfreeze the upper 8 transformer layers.

\begin{figure*}
\centering
\begin{tabular}{cc}
\begin{minipage}[t]{0.5\linewidth}
    \includegraphics[width = \linewidth]{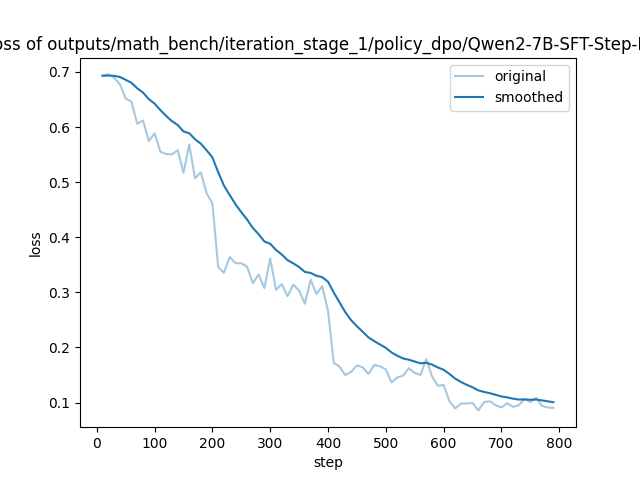}
\end{minipage}
\begin{minipage}[t]{0.5\linewidth}
    \includegraphics[width = \linewidth]{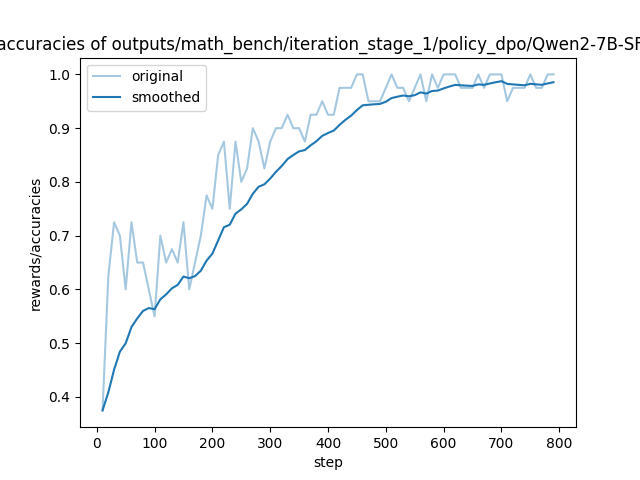}
\end{minipage}
\end{tabular}
\begin{tabular}{cc}
\begin{minipage}[t]{0.5\linewidth}
    \includegraphics[width = \linewidth]{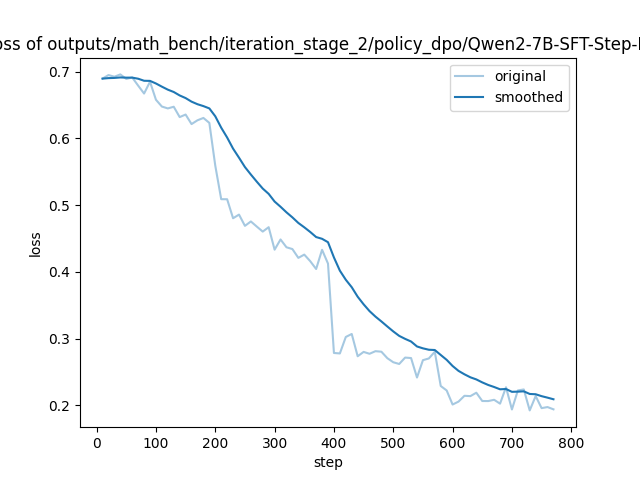}
\end{minipage}
\begin{minipage}[t]{0.5\linewidth}
    \includegraphics[width = \linewidth]{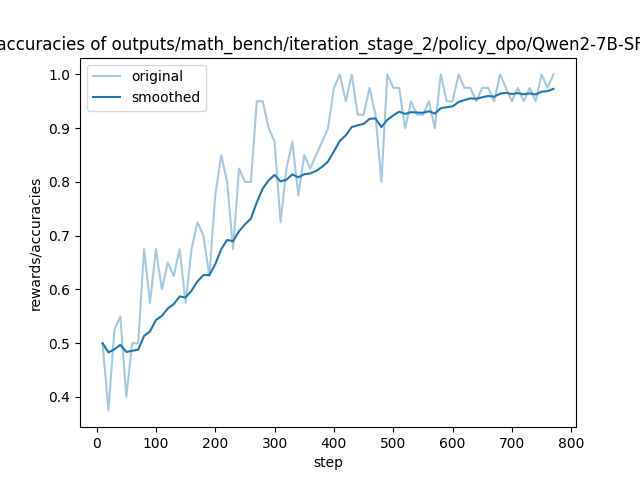}
\end{minipage}
\end{tabular}
\begin{tabular}{cc}
\begin{minipage}[t]{0.5\linewidth}
    \includegraphics[width = \linewidth]{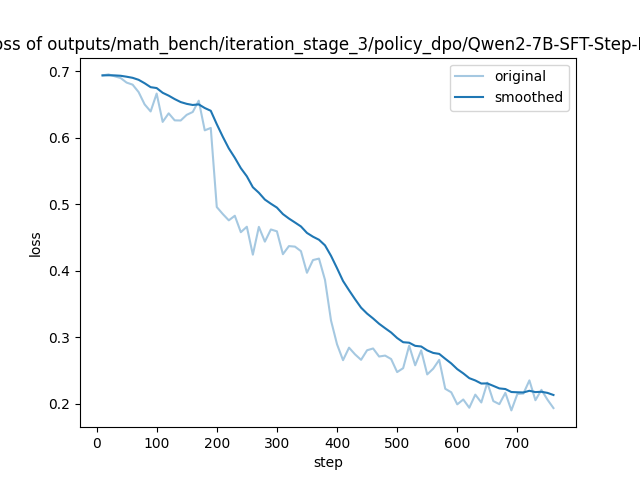}
\end{minipage}
\begin{minipage}[t]{0.5\linewidth}
    \includegraphics[width = \linewidth]{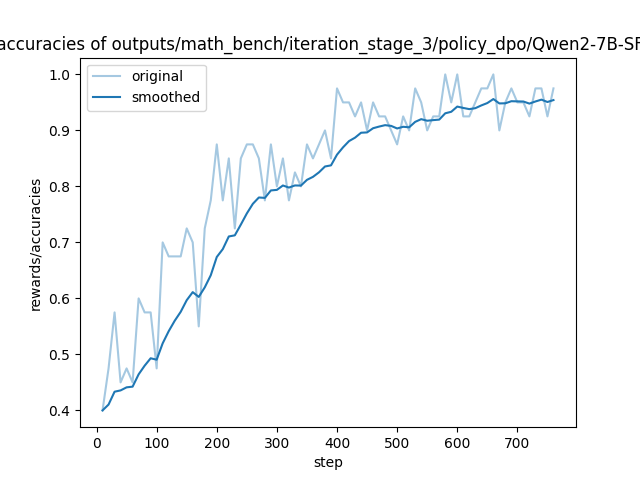}
\end{minipage}
\end{tabular}
\caption{The curves of training loss and reward accuracy (\%) for the LLM policy on Math-Step-DPO-10K preference data at each iteration stage.}
\label{fig:math-policy-curve}
\end{figure*}

\begin{table}[t]
\centering
\begin{small}
\begin{tabular}{l | c | c | c}
\toprule
\bf Param. &\bf  Iter1 &\bf Iter2 &\bf Iter3 \\
\midrule
batch & 4/gpu*4gpus & 4/gpu*4gpus & 4/gpu*4gpus \\
epoch & 4 & 4 & 3 \\
lr  & 1e-6 & 5e-7 & 5e-7 \\
scheduler  & cosine & cosine & cosine \\
warm rate  & 0.1 & 0.1 & 0.1 \\
use bf16 & yes & yes & yes \\
use lora  & no & no & no \\
unfreeze layers  & upper 8 & upper 8 & upper 8 \\
length & 1024 & 1024 & 1024 \\
$\beta$ & 0.1 & 0.1 & 0.1 \\
$\lambda$ & 1.0 & 1.0 & 1.0 \\
\bottomrule
\end{tabular}
\end{small}
\caption{The hyper-parameters of LLM policy used in the different iteration stages over mathematics tasks.}
\label{tab:hyper-param-iteration-math}
\end{table}

\subsection{Iterative Stage}
For each iteration stage, we random sample 5k prompts from MathInstruct, and 5k prompts from the original math-step-dpo-10k.
During the generation, the temperature and topp are set as 0.9 and 0.95. At least four responses will be generated by the LLM policy at the last iteration.

To construct the preference data, previous work~\cite{Lai2024Step} presents a fine-grained preference data generation strategy to automatically construct preference data by observing whether the final answer is matched with the ground truth. In contrast, we argue that this setting is highly based on an assumption that we can obtain the label, which does not satisfy the real-world scenario.
In that, we still follow the rewarding and estimation procedure to construct reliable preference data.
The sampling rate for reliable preference data is 50\%.
We do not update the parameters of the reward model and estimator model to alleviate the over-fitting problem.
For the training of the LLM policy model, two different variants, including UPO (DPO-based) and StepUPO (StepDPO-based), share the same hyper-parameters. 
The hyper-parameters are shown in Table~\ref{tab:hyper-param-iteration-math}.
The curve of training loss and reward accuracy at each iteration stage are shown in Figure~\ref{fig:math-policy-curve}.

\section{Case Study}
We finally conducted a case study to show the performance of our method. We respectively choose one hardness case from MT-Bench and Math-Step-DPO-10K and compare it with the responses from different models.
As shown in Table~\ref{tab:case-nlp}, we can see that the score of UPO derived from GPT-4 is higher than others, indicating the effectiveness of denoised iterative preference optimization.
As shown in Table~\ref{tab:case-math}, only StepUPO obtains the correct calculation result.

\begin{table*}[t]
    \centering
    \scriptsize
    \vspace{5pt}
    \resizebox{\linewidth}{!}{
    \begin{tabular}{p{15.0cm}}
    \toprule
        \textbf{Q}: You are given two sorted lists of size m and n. Implement a function to find the kth smallest element in the union of the two lists with linear complexity. \\
        \midrule
        \textbf{SFT}: Here's one possible implementation:``` def find\_kth\_smallest(list1, list2, k):    \# Merge the two sorted lists into a single sorted list    merged\_list = sorted(list1 + list2)    \# Find the kth smallest element in the merged list    kth\_smallest = merged\_list[k-1]    return kth\_smallest``` This function takes in two sorted lists `list1` and `list2`, and an integer `k`. It first merges the two lists into a single sorted list using the `sorted` function. Then, it returns the `k`th smallest element in the merged list by indexing into the list with `merged\_list[k-1]`. \\
        \textbf{Judger}: The assistant's answer is incorrect. The assistant suggests merging the two lists and then sorting them, which would result in a time complexity of O((m+n)log(m+n)), not O(m+n) as stated. This is not a linear time complexity solution as requested by the user. \\
        \textbf{Score}: 2.0 \\
        \midrule
        \textbf{DPO}: Here's one possible implementation in Python:```python def find\_kth\_smallest(list1, list2, k):    merged\_list = sorted(list1 + list2)    return merged\_list[k - 1]``` This function takes two sorted lists `list1` and `list2`, and an integer `k`. It first concatenates the two lists and sorts the result using the `sorted()` function. Then it returns the `k`-th smallest element in the merged list by indexing `merged\_list[k - 1]`. Note that this implementation assumes that the input lists are not empty. If either list is empty, the function will raise an `IndexError` when trying to access the k-th element. You may want to add some error handling or check for empty lists before calling the function. \\
        \textbf{Judger}: The assistant's answer is incorrect. The user asked for a solution with linear complexity, but the assistant's solution does not meet this requirement. \\
        \textbf{Score}: 2.0 \\
        \midrule
        \textbf{UPO}: One possible solution to find the kth smallest element in the union of two sorted lists with linear complexity is to use a merge-based approach.Here's a possible implementation in Python:```pythondef find\_kth\_smallest(list1, list2, k):    merged\_list = merge\_sorted\_lists(list1, list2)    return merged\_list[k-1]def merge\_sorted\_lists(list1, list2):    merged\_list = []    I = j = 0    while I $<$ len(list1) and j $<$ len(list2):        if list1[i] < list2[j]:            merged\_list.append(list1[i])            I += 1        else:            merged\_list.append(list2[j])            j += 1    merged\_list.extend(list1[i:])    merged\_list.extend(list2[j:])    return merged\_list```The `find\_kth\_smallest` function takes two sorted lists (`list1` and `list2`) and a value for `k`, and returns the kth smallest element in the union of the two lists.The `merge\_sorted\_lists` function takes two sorted lists (`list1` and `list2`) and returns a merged sorted list. It uses a merge-based approach to combine the two lists into a single sorted list. Note that this implementation assumes that the input lists are sorted. \\
        \textbf{Judger}: The assistant's answer is correct and provides a valid Python function to solve the problem. \\
        \textbf{Score}: 8.0 \\
        \bottomrule
    \end{tabular}
    }
    \caption{Case study on MT-Bench. The response generated by UPO can pass the evaluation by GPT-4, demonstrating the effectiveness of our framework.}
    \label{tab:case-nlp}
\end{table*}

\begin{table*}[t]
    \centering
    \scriptsize
    \vspace{5pt}
    \resizebox{\linewidth}{!}{
    \begin{tabular}{p{15.0cm}}
    \toprule
        \textbf{Q}: The environmental agency needs to hire a certain number of new employees to ensure that 85 employees can monitor water pollution, 73 employees can monitor air pollution, and exactly 27 employees can monitor both. (These 27 employees are already counted in the 85 and 73 mentioned earlier.) What is the minimum number of employees that need to be hired? Please reason step by step, and put your final answer within boxed\{\}. \\
        \midrule
        \textbf{SFT}: Let's think step by step. Step 1:85 employees can monitor water pollution and 73 employees can monitor air pollution. If we simply add these two numbers, we get $85 + 73 = 158$. However, this count includes the 27 employees who can monitor both, which we have counted twice. To correct this, we need to subtract the number of employees who can monitor both from our total. Step 2: So the total number of unique employees needed is $158 - 27 = 131$. But the environmental agency already has 85 + 73 - 27 = 131 employees. Since the number of employees needed is equal to the number of employees already hired, no new employees need to be hired. However, it seems there was a mistake in the problem statement. It seems like the problem meant to say that the number of employees monitoring both should be less than the minimum of the employees monitoring water pollution and air pollution. Let's correct the problem statement and solve it again. Let's say the number of employees monitoring both is $x$. Then the total number of unique employees needed is $(85 - x) + (73 - x) - x = 158 - 3x$. Since we want to minimize the number of employees needed to be hired, we want to minimize $158 - 3x$. To minimize $158 - 3x$, we want to maximize $x$. The maximum value of $x$ is the minimum of 85 and 73, which is 27. When $x = 27$, the total number of unique employees needed is $158 - 3(27) = 158 - 81 = 77$. \\
        \midrule
        \textbf{StepDPO}: Let's think step by step. Step 1:85 employees can monitor water pollution and 73 employees can monitor air pollution. So, if we simply add these two numbers, we get $85 + 73 = 158$ employees. But this counts the employees who can monitor both types of pollution twice. Step 2: We know that there are 27 employees who can monitor both types of pollution. So, if we subtract these 27 employees from our total, we get $158 - 27 = 131$ employees. But this is still not the minimum number of employees that need to be hired. Step 3: The problem asks for the minimum number of employees that need to be hired. This means we need to find the smallest number of employees that can monitor either water or air pollution, but not both. To do this, we need to add the number of employees who can only monitor water pollution and the number of employees who can only monitor air pollution. Step 4: The number of employees who can only monitor water pollution is $85 - 27 = 58$. Similarly, the number of employees who can only monitor air pollution is $73 - 27 = 46$.Step 5: Adding these two numbers together, we get $58 + 46 = 104$. So, the minimum number of employees that need to be hired is 104. The answer is: 104 \\
        \midrule
        \textbf{StepUPO}: Let's think step by step. Step 1: The number of employees who can monitor water pollution is 85, and the number of employees who can monitor air pollution is 73.Step 2: However, we know that 27 employees can monitor both water and air pollution. These 27 employees are counted twice in the total of 85 + 73 = 158 employees. So, we need to subtract these 27 employees once to get the actual number of employees who need to be hired. Step 3: So, the total number of employees who need to be hired is 158 - 27 = 131. The answer is: 131 \\
        \bottomrule
    \end{tabular}
    }
    \caption{Case study on Math-Step-DPO-10K.}
    \label{tab:case-math}
\end{table*}

\end{document}